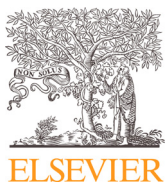
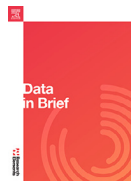

Data Article

# ArzEn-MultiGenre: An aligned parallel dataset of Egyptian Arabic song lyrics, novels, and subtitles, with English translations

Rania Al-Sabbagh

*Department of Foreign Languages, University of Sharjah, United Arab Emirates*

## ABSTRACT

ArzEn-MultiGenre is a parallel dataset of Egyptian Arabic song lyrics, novels, and TV show subtitles that are manually translated and aligned with their English counterparts. The dataset contains 25,557 segment pairs that can be used to benchmark new machine translation models, fine-tune large language models in few-shot settings, and adapt commercial machine translation applications such as Google Translate. Additionally, the dataset is a valuable resource for research in various disciplines, including translation studies, cross-linguistic analysis, and lexical semantics. The dataset can also serve pedagogical purposes by training translation students and aid professional translators as a translation memory. The contributions are twofold: first, the dataset features textual genres not found in existing parallel Egyptian Arabic and English datasets, and second, it is a gold-standard dataset that has been translated and aligned by human experts.

© 2024 The Author(s). Published by Elsevier Inc.
This is an open access article under the CC BY license
(http://creativecommons.org/licenses/by/4.0/)

Specifications Table

| | |
|---|---|
| Subject | Computer Science, Social Sciences |
| Specific subject area | Natural Language Processing, machine translation, large-language models, translation studies, cross-linguistic analysis, lexical semantics |
| Data format | Translated and aligned |
| Type of data | Texts (Bilingual tables in Microsoft Excel files) |
| Data collection | The ArzEn-MultiGenre dataset consists of three genres: song lyrics, novels, and subtitles. The data was gathered from various sources using different methods. A website was crawled for song lyrics using an in-house web crawler, and professional translators manually translated the lyrics into English. For novels, hard copies were collected in English and Egyptian Arabic, then scanned and converted into text files using an Optical Character Recognizer (OCR). The OCR output was then manually reviewed and aligned. Finally, Egyptian Arabic and English subtitles were manually transcribed from two Netflix series and then aligned. |
| Data source location | The data used was obtained from various sources through various methods:<br>- Aghani Lyrics, a website, was used to collect Egyptian Arabic song lyrics<br>- BeautifulSoup4 v.4.12.2, a Python library, was used to parse the collected song lyrics.<br>- The Egyptian Arabic translations of the novels were purchased at Hunna, a publishing company in Egypt.<br>- The English novels were purchased from Amazon UK.<br>- Sotoor, an AI optical character reader, converted the hard copies of the English and Egyptian Arabic novels into text files.<br>- The TV shows were collected from Netflix.<br>- Professional translators (to translate the song lyrics) and language experts (to proofread the OCR output, transcribe the subtitles, and align the segments) were hired in Egypt. |
| Data accessibility | Repository name: Mendeley Data<br>Data identification number: 10.17632/6k97jty9xg.4<br>Direct URL to data: https://data.mendeley.com/datasets/6k97jty9xg/4. |

## 1. Value of the Data

- ArzEn-MultiGenre [1] includes three textual genres unrepresented in current parallel Egyptian Arabic and English datasets. These genres are song lyrics, novels, and subtitles. Current datasets feature genres such as personal interviews [2], weblogs [3], SMS and chat messages [4], discussion forums [5], travel questions and answers [6], and unscripted telephone conversations [7,8]. While the Habibi corpus [9] includes song lyrics from six Arabic dialects, including Egyptian Arabic, it remains monolingual. The OPUS subtitles corpus [10] provides thousands of English and Arabic parallel subtitles, but the Arabic subtitles are exclusively in Modern Standard Arabic, not Egyptian Arabic.
- ArzEn-MultiGenre is entirely translated and aligned manually, making it a gold-standard resource. The manual effort distinguishes ArzEn-MultiGenre from current parallel datasets that rely on crowdsourced translation, such as [3] and [8].
- ArzEn-MultiGenre comprises 25,557 segment pairs compared to 2,600 from Zbib et al. [3] and 12,000 from Bouamor et al. [6]. Kumar et al. [8] have 35,892 segment pairs; however, these were translated by crowdsourced amateur translators, unlike the segments of ArzEn-MultiGenre that were translated by professional translators (i.e., translators with formal training in translation, extensive experience with paid translation services, complete mastery of the source and target languages, and full knowledge of the cultural differences between the source and target audiences).
- ArzEn-MultiGenre is a gold-standard dataset that natural language processing researchers can use to evaluate the performance of new machine translation models across different genres. It is also suitable for fine-tuning large language models, especially in few-shot settings. The

Table 1

BLEU scores when using ArzEn-MultiGenre to customize Google Translate using AutoML Translation.

| Genre | BLEU Scores | | Dataset Stats | | |
| --- | --- | --- | --- | --- | --- |
| | Google NMT | Customized Model | Training | Validation | Test |
| Songs | 8.95 | 11.87 | 2439 | 304 | 304 |
| Novels | 11.84 | 17.28 | 4151 | 518 | 518 |
| Subtitles | 12.54 | 19.05 | 6755 | 844 | 844 |

*Note:* Google NMT refers to Google's neural machine translation original model, and the customized model refers to the customized model using AutoML Translation. The training, validation, and test refer to the number of pairs for each process.

- genres represented in ArzEn-MultiGenre present challenges for automated models. They include numerous one-word segments, idiomatic expressions, culture-specific references, and slang words. Moreover, dialects, evolving more rapidly than standard varieties, introduce ongoing linguistic shifts, including the constant creation and borrowing of new words and frequent semantic transformations. Therefore, it is crucial to continuously train and evaluate automated models using up-to-date data to ensure their effectiveness in handling such complexities.
- Researchers in translation studies can benefit from ArzEn-MultiGenre in several ways. For example, it can be used to compare translation strategies across genres: how translating figurative language and cultural references may differ between translating novels, where the translator has more space to write an explanation, and subtitles, where there is only a limited number of characters for each subtitle. Also, it can help researchers study how humans translate gender in song lyrics. This is especially important since Arabic is a heavily gendered language, and Egyptian Arabic, in particular, uses the masculine gender to refer to female lovers. It is worth noting that gender in Arabic machine translation is still a major issue that needs to be addressed [11,12].
- ArzEn-MultiGenre ArzEn-MultiGenre is useful for professional translators and translation students. Video-on-demand streaming services such as Netflix subtitle Egyptian Arabic series and movies into English, and there is a tendency to translate world literature into Egyptian Arabic rather than the formal variety of Modern Standard Arabic. Therefore, ArzEn-MultiGenre can help train translation students and function as a translation memory for professional translators. ArzEn-MultiGenre also allows professionals and students to customize commercial machine translation systems such as Google Translate to suit their needs better. Using ArzEn-MultiGenre to customize Google Translate through AutoML Translation led to higher BLEU scores (see Table 1).

## 2. Background

Several datasets are available for Arabic and English, but most are limited to standard Arabic, which is used for formal written and spoken communication. Datasets for dialectal Arabic and English are fewer and smaller. Therefore, ArzEn-MultiGenre was created to increase the number of parallel datasets for Arabic dialects and English. Moreover, as mentioned in the previous section, this dataset covers various textual genres not included in the existing Egyptian Arabic and English datasets. Recently, using the song lyrics from ArzEn-MultiGenre, I demonstrated that training machine translation applications such as Google Translate on too many and too large standard Arabic-English parallel datasets and too few dialectal Arabic-English parallel datasets had a negative transfer impact [13].

In this data article, ArzEn-MultiGenre is introduced in detail, along with two new parts that were added, namely novels and subtitles. Currently, I am benchmarking AI applications such as Bard and ChatGPT for translating Egyptian Arabic subtitles. The focus is on how these applications translate one-word discourse fillers, cultural references, idioms, and figurative language.

Additionally, I want to compare the performance of these AI applications with Google Translate and assess how much post-editing effort they can save.

## 3. Data Description

The ArzEn-MultiGenre repository has three folders: songs, novels, and subtitles. The songs folder contains one file with all song lyrics together. The first column is the song's name, the second is the year of the album release, the third is the album's name, the fourth is the song number in the dataset, the fifth is the Egyptian Arabic lyrics, and the sixth is the English translations. There is a total of 280 song lyrics. The novels folder contains three files, one file per novel. Finally, the subtitles folder contains 12 files, one file for each episode. The files in the novels and subtitles folders are formatted as bilingual tables, where each row includes a pair of source and target segments. The first column of the table is for the source language, whereas the second column is used for the target language. The files are available in Microsoft Excel formats.

## 4. Experimental Design, Materials and Methods

### 4.1. Compilation and pre-processing

Different methods were used to collect and pre-process each part of ArzEn-MultiGenre, as explained in subsections 4.2–4.4.

### 4.2. ArzEn-MultiGenre – novels

Recently, three novels have been translated into Egyptian Arabic. These include Ernest Hemingway's *The Old Man and the Sea*, Antoine Saint-Exupéry's *The Little Prince*, and Albert Camus's *The Stranger*. Magdy Abdelhadi translated the first, while Hector Fahmy translated the last two. Both translators work for an Egyptian publishing company called Hunna and are certified translators in Egypt. *The Old Man and the Sea* explores themes of man versus nature, courage and endurance, the heroic ideal, mortality and death, and the quest for meaning, all woven into a poignant narrative of human resilience and the pursuit of purpose. *The Little Prince* discusses several themes, including the importance of imagination and creativity, the value of human connections and friendship, the pursuit of truth and meaning in life, and the contrast between childhood innocence and adult perception. *The Stranger* delves into themes of existentialism, absurdity, and the universe's indifference.

The novels were available only as hard copies. Therefore, after purchasing the original novels and the translations, the hard copies were scanned, and Sotoor [14] converted them into text files. Sotoor is an optical character reader that supports Arabic and has an API option for batch processing. Proofreaders were hired to manually check the accuracy of the converted texts. The proofreaders were native speakers of Egyptian Arabic and graduates of English language departments in Egyptian universities.

The proofreading instructions were to maintain the spelling and punctuation of the source and target texts. Like most other Arabic dialects, Egyptian Arabic lacks specific spelling and punctuation rules. People tend to write based on their pronunciation, which can vary greatly depending on their educational and social background. Similarly, there are no specific rules governing the use of punctuation marks. To accurately reflect these features of Egyptian Arabic, I requested that proofreaders adhere to the actual spelling and punctuation used in the hard copies and not impose their own rules.

**Table 2**

Examples of different translations of the same source segment by Netflix.

| Arz Script | En SDH | En non-SDH |
| --- | --- | --- |
| طب ما بلاش النهاردة يا رئيفة | Why not on another day, Raeefa? | Let's skip my birthday today, Raeefa. |
| أنت عارفة أنا بأكره أعياد الميلاد والمفاجئات والتورتة والشموع | You know how much I hate birthdays, surprises, cakes, and candles. | You know I hate all that. And I don't want anyone to make a big deal. Just take the cake back. |
| عيد ميلاد إيه بس؟ | What birthday? | What birthday? |
| ده ولا في دماغي. | I'm not even thinking about that. | It never even crossed my mind. |
| ولا في تورتة ولا في حاجة، قعدة عادية كده هوه | There is no cake and whatnot. It is just a normal gathering. | I didn't buy a cake. It's just lunch. |
| إلهام بس كان نفسها في شوية ملوخية قولت بقا أعملهالها وأعمل معاها إيه الكفتة اللي أنت بتحبها | Elham craved mulukhia, so I figured I'd cook it with the kufta you love. | Elham was craving Mulukhiyah. I decided to make some with the meatballs that you like. |

### 4.3. ArzEn-MultiGenre – subtitles

In recent years, video-on-demand streaming platforms like Netflix have begun producing Egyptian Arabic series and providing English subtitles. For instance, *Paranormal* and *Finding Ola* are two shows that were released in November 2020 and February 2022, respectively. *Paranormal* follows the life of Refaat Ismail, a hematologist grappling with supernatural phenomena and unexplainable events. Set in the 1960s in Egypt, the show blends mystery, suspense, and the supernatural as Refaat navigates through encounters with ghosts, demons, and otherworldly forces. Through his investigations, Refaat confronts his skepticism and delves into the mysteries of the unknown, all while unraveling secrets from his past. The series blends Egyptian folklore, history, and contemporary storytelling. In *Finding Ola*, a divorced Egyptian mother defies social expectations to rediscover herself. Juggling motherhood, work, and self-discovery, she navigates societal pressures, family dynamics, and the search for love and joy, all while offering a glimpse into contemporary Egyptian life and celebrating female empowerment through her journey.

Netflix has different subtitling options. One option is the Egyptian Arabic subtitles, primarily for those who are deaf or hard of hearing (SDH). These subtitles include the dialogue and information about the sound effects and music. English subtitles are also available in two versions - an SDH version and a non-SDH version with only the dialogue. The differences between the English SDH and non-SDH versions are not only in the presence or absence of non-dialogic information but there are differences in the English translation of the dialogue itself. Consider Table 2 to see how the same Egyptian Arabic segments from a *Paranormal* episode have two completely different translations.

Arabic transcribers fluent in English were employed to transcribe the subtitles in text files according to the following guidelines:

- For the Egyptian Arabic subtitles, information other than the dialogue should be excluded.
- The SDH and non-SDH subtitles English subtitles are to be transcribed, but non-dialogic information such as sound effects and music should be excluded.
- Egyptian Arabic should be written exactly as shown on the screen, with no changes to spelling or punctuation, except when codemixing occurs.

  *Paranormal* and *Finding Ola* showcase well-educated and upper-class Egyptians who frequently mix English with Egyptian Arabic at both the word and sentence levels. There are two forms of word-level codemixing: saying the entire word in English or combining Egyptian Arabic affixes with an English word. For instance, "biyi-flirt" means "he's flirting" and is formed by attaching the English verb "flirt" to two Egyptian Arabic prefixes: "bi-" for the progressive aspect and "yi" for the third-person masculine present tense.

The transcribers were given instructions to avoid including instances of sentence-level codemixing. This means that if a sentence is spoken entirely in English, it should not be written down. However, in cases where there is word-level codemixing, the transcribers had two options. If the English word is commonly written using Arabic script, such as "okay" (which is commonly written as اوكيه, اوك, or اوكي) or "merci" (which is commonly written as ميرسي or مرسي), then it should be written in Arabic script using any spelling that the transcribers choose. However, if the English word is not typically written in Arabic script, it should be written either in English (if it was said entirely in English) or in a mix of English and Arabic (if it has an Arabic component). For example, the word "biyi-flirt" was transcribed as (بيflirt) in the dataset.

### 4.4. ArzEn-MultiGenre – songs lyrics

An Arabic website called Alaghani Lyrics [15] includes many Arabic song lyrics written by fans. Searching the website for Amr Diab's songs, 280 songs were found, ranging from 1983 to 2023. Amr Diab's songs are exclusively in Egyptian Arabic, i.e., they are neither mixed with English nor Modern Standard Arabic. Common themes in Amr Diab's songs often revolve around love, romance, and relationships, portraying the complexities of human emotions and experiences. His lyrics frequently explore the exhilaration of falling in love, the longing for a beloved, and the pain of heartbreak. Diab's music also frequently touches upon themes of nostalgia, celebrating memories of the past and the passage of time. Additionally, themes of perseverance, resilience, and optimism often emerge, conveying hope and strength in the face of life's challenges.

An in-house web crawler was used to scrap the HTML files of the songs, and then the files were cleaned using Beautiful Soup 4.1. [16] Duplicate lines were removed. The scraped lyrics were in Egyptian Arabic ; therefore, four translators and one reviewer were hired to translate the lyrics. All translators and the reviewer were native speakers of Egyptian Arabic and well acquainted with the songs. The translators each had four years, and the reviewer had six years of experience with Egyptian Arabic–English subtitling for Netflix. The corpus was divided equally among the four translators, and then the reviewer checked the translations for accuracy and consistency. The translation guidelines were as follows:

- Translators must provide natural translations that reflect the intended meaning.
- Translators must avoid literal translations of idioms, figures of speech, and culture-specific references.
- Translators must use mainstream English without slang.
- Translators must maintain American English spelling.
- Translators should ignore singability in translations, i.e., translations need not follow rhyme or rhythm patterns.
- Translators should account for blurbs, such as repetition, in their translations.
- Translators should keep the punctuation marks of the source text in the same relative position.
- Translators must not use machine translation or artificial intelligence applications; sentences must be translated manually from scratch.
- Translators should listen to the songs to decipher ambiguous words.

## 5. Segmentation

In preparation for alignment, the source and target texts were segmented so that each segment ends with a major punctuation mark (i.e., a period, exclamation, or question mark) or a new-line break. Some segments were stand-alone utterances, mainly discourse fillers such as segments 2 and 4 in Table 3. Others were stand-alone phrases, such as segment 3 in Table 3. Also, some segments were stand-alone simple, complex, and compound sentences (segments 1, 5–13 in Table 3).

**Table 3**
A sample of ArzEn-MultiGenre subtitles to show different types of segments.

| No. | Source Text | Translation |
|---|---|---|
| 1 | أنا مبفكرش في عيالي؟ | I don't think of the kids. Me? |
| 2 | أيوه | Yes |
| 3 | أنا؟ | Me? |
| 4 | ايوه | Yes. |
| 5 | طب أنا ببقى مشغول طول النهار معرفش تفاصيلهم، أنت قاعدة معاهم بتعملي إيه؟ | I work all day, and I don't follow them. You're at home doing what? |
| 6 | بترسمي السيناريو المثالي لكل حاجة. | Planning the perfect scenario for everything. |
| 7 | ولو حد خالف توقعاتك أو خرج برة النص، تلومي كل الناس ماعدا نفسك، صح؟ | When someone doesn't play by your rules, you blame everybody else but yourself. |
| 8 | أنت شخصية توكسيك | You're toxic. |

## 6. Alignment

Four annotators were hired to align the source and target segments. The annotators were native Arabic speakers who were well acquainted with Egyptian Arabic and graduates of English departments in Egypt. The alignment was done using Microsoft Excel.

Annotators were instructed that there are different correspondences between the source and target segments:

a. One-to-one: One source segment is translated into one target segment.
b. One-to-many: One source segment is translated into multiple segments. In this case, the target segments must be merged into one cell.
c. Many-to-one: Multiple source segments are translated into one segment. In this case, the source segments must be merged into one cell.
d. One/many-to-zero: One or more source segments are untranslated; therefore, the target segment cell should be left empty.
e. Zero-to-one/many: There is a target segment that maps to no source segment. This is usually the case when translators add more explanation in English to clarify an idiom, a figure of speech, or a culture-specific reference.

The corpus was equally divided across the annotators. Once each annotator finished their part, they exchanged the parts to review each other's work. Furthermore, I reviewed all alignments as an adjudicator.

## 7. Results

Table 4 shows the statistics of ArzEn-MultiGenre in terms of word tokens, types, and type-token ratio. Table 5 shows the different lengths of segments in each genre. Subtitles had 483 segments with zero correspondences (either the source or the target segment was missing). Novels had 32 segments with zero correspondences. However, no zero correspondences were found in the translations of the song lyrics because translators were instructed to translate all segments.

**Table 4**
Statistics of ArzEn-MultiGenre.

| Genre | Segment Pairs | Word Tokens | | Word Types | | Type-Token Ratio | |
|---|---|---|---|---|---|---|---|
| | | Arz | En | Arz | En | Arz | En |
| Novels | 5226 | 56,527 | 78,890 | 15,590 | 10,246 | 28% | 13% |
| Subtitles | 17,265 | 80,273 | 99,884 | 12,798 | 10,032 | 16% | 10% |
| Songs | 3066 | 17,858 | 31,294 | 5680 | 3719 | 32% | 12% |
| **Totals** | **25,557** | **154,658** | **210,068** | **29,179** | **18,131** | **19%** | **9%** |

**Table 5**
Segment lengths in ArzEn-MultiGenre.

| Genre | Arz segment lengths | | | En segment lengths | | |
|---|---|---|---|---|---|---|
| | 1 word | 2–5 words | 6+ words | 1 word | 2–5 words | 6+ words |
| Novels | 54 | 1269 | 3903 | 36 | 618 | 4571 |
| Songs | 17 | 1556 | 1493 | 1 | 414 | 2651 |
| Subtitles | 2689 | 9252 | 5324 | 1920 | 8202 | 7143 |
| **Totals** | **2744** | **12,077** | **10,720** | **1957** | **9234** | **14,365** |

**Limitations**

The two novels, *The Little Prince* by Antoine Saint-Exupéry and *The Stranger* by Albert Camus, were originally written in French. In ArzEn-MultiGenre, I used the English translations of these novels. Specifically, I used Katherine Woods's translation of *The Little Prince* and Stuart Gilbert's translation of *The Stranger*. It is worth noting that Hector Fahmy translated the Egyptian Arabic versions of these novels directly from French rather than English.

It is important to note that when using Google AutoML Translation or Cloud Translation with ArzEn-MultiGenre, Google automatically removes segment pairs with the same source segment, even if the target segment is different. This is why there are fewer segment pairs in Table 1 than in Tables 4 and 5.

**Ethics Statement**

The author has read and followed the ethical requirements for publication in *Data in Brief* and confirms that the current work does not involve human subjects, animal experiments, or any data collected from social media platforms.

**CRediT Author Statement**

This is a single-authored article.

**Data Availability**

ArzEn-MultiGenre (Original data) (Mendeley Data).

**Acknowledgements**

This research did not receive any specific grant from funding agencies in the public, commercial, or not-for-profit sectors.

## Declaration of Competing Interest

The author declares that she has no known competing financial interests or personal relationships that could have appeared to influence the work reported in this paper.